\documentclass[sigconf]{acmart}

\AtBeginDocument{%
  }

\copyrightyear{2025}
\acmYear{2025}
\setcopyright{acmlicensed}
\acmConference[CIKM '25]{Proceedings of the 34th ACM International Conference on Information and Knowledge Management}{November 10--14, 2025}{Seoul, Republic of Korea}
\acmBooktitle{Proceedings of the 34th ACM International Conference on Information and Knowledge Management (CIKM '25), November 10--14, 2025, Seoul, Republic of Korea}
\acmDOI{10.1145/3746252.3761052}
\acmISBN{979-8-4007-2040-6/2025/11}
\usepackage{soul}
\usepackage{url}
\usepackage{amsmath}
\usepackage{amsthm}
\usepackage{booktabs}
\usepackage{algorithm}
\usepackage{algorithmic}
\usepackage[switch]{lineno}
\usepackage{multirow}
\usepackage{colortbl}
\usepackage{enumitem}
\usepackage{tikz}
\usepackage{pythonhighlight}
\usepackage{xcolor}
\usepackage{tablefootnote}

\begin{document}

\title{Multi-Turn Interactions for Text-to-SQL with Large Language Models}

\author{Guanming Xiong}
\orcid{0000-0001-8634-3669}
\affiliation{
  \institution{Peking University}
  \city{Beijing}
  \country{China}
}
\email{gm\_xiong@pku.edu.cn}

\author{Junwei Bao}
\orcid{0000-0002-5549-5130}
\authornote{Corresponding author.}
\affiliation{
  \institution{Zuoyebang Education Technology Co., Ltd.}
  \city{Beijing}
  \country{China}
}
\email{baojunwei001@gmail.com}

\author{Hongfei Jiang}
\orcid{0009-0001-0553-4568}
\affiliation{
  \institution{Zuoyebang Education Technology Co., Ltd.}
  \city{Beijing}
  \country{China}
}
\email{jianghongfei@zuoyebang.com}

\author{Yang Song}
\orcid{0009-0008-4045-031X}
\affiliation{
  \institution{Zuoyebang Education Technology Co., Ltd.}
  \city{Beijing}
  \country{China}
}
\email{songyang@zuoyebang.com}

\author{Wen Zhao}
\orcid{0000-0002-5760-4759}
\affiliation{
  \institution{Peking University}
  \city{Beijing}
  \country{China}
}
\email{zhaowen@pku.edu.cn}
\renewcommand{\shortauthors}{Guanming Xiong, Junwei Bao, Hongfei Jiang, Yang Song, and Wen Zhao}

\begin{abstract}
This study explores text-to-SQL parsing by leveraging the powerful reasoning capabilities of large language models (LLMs). 
Despite recent advancements, existing LLM-based methods are still inefficient and struggle to handle cases with wide tables effectively.
Furthermore, current interaction-based approaches either lack a step-by-step, interpretable SQL generation process or fail to provide a universally applicable interaction design.
To address these challenges, we introduce Interactive-T2S, a framework that generates SQL queries through direct interactions with databases. 
This framework includes four general tools that facilitate proactive and efficient information retrieval by the LLM. 
Additionally, we have developed detailed exemplars to demonstrate the step-wise reasoning processes within our framework. 
Our approach achieves advanced performance on the Spider and BIRD datasets as well as their variants. Notably, we obtain state-of-the-art results on the BIRD leaderboard under the setting without oracle knowledge, demonstrating the effectiveness of our method.\footnote{Code and data are available at: \url{https://github.com/JimXiongGM/Interactive-Text-to-SQL}}
\end{abstract}



\maketitle
\section{Introduction}

Text-to-SQL technology, which translates natural language questions into executable SQL queries, has emerged as a crucial field of research. This technology empowers non-experts to interact with relational databases (DBs), which have become ubiquitous in the era of big data \cite{Hong-Zijin-2024-ArXiv-Survey-LLM-based-Text-to-SQL}. A significant challenge in this field is designing a text-to-SQL system that operates accurately and efficiently within resource constraints.

The emergence of large language models (LLMs), such as ChatGPT \cite{Ouyang-Long-NeurIPS-2022-InstructGPT} and GPT-4 \cite{OpenAI-2023-GPT4}, has opened new avenues for enhancing text-to-SQL systems. These models have shown promising capabilities in reasoning and few-shot learning, establishing new benchmarks in this domain \cite{Gao-Dawei-VLDB-2024-DAIL-SQL,Pourreza-Mohammadreza-NeurIPS-2023-DIN-SQL}.

Recent advancements in text-to-SQL research encompass two primary perspectives: prompt optimization and interaction strategies \cite{Hong-Zijin-2024-ArXiv-Survey-LLM-based-Text-to-SQL}. Prompt optimization focuses on crafting prompts that guide LLMs to generate accurate SQL queries. This involves constructing precise schema linking, leveraging similar examples, and employing effective question decomposition methods \cite{Gao-Dawei-VLDB-2024-DAIL-SQL,Zhang-Hanchong-EMNLP-2023-ACT-SQL}. Interaction strategies, on the other hand, center around designing methods to refine SQL queries through execution-based feedback \cite{Qu-Ge-2024-ArXiv-TA-SQL,Shi-Freda-EMNLP-2022-MBR-Exec,Chen-Xinyun-ICLR-2024-SELF-DEBUGGING}. Recent approaches also introduce interactive models that leverage specific tools to interact with DBs, yielding significant improvements \cite{Jiang-EMNLP-2023-StructGPT,Gu-Yu-2024-ArXiv-FUXI}.

\begin{figure}
    \centering
    \includegraphics[width=0.33\textwidth,page=1]{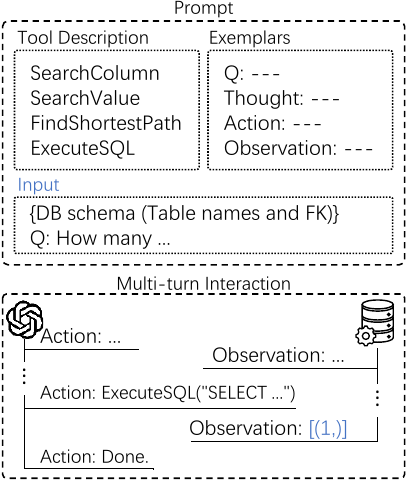}
    \caption{Overview of the interactive process.}
    \label{fig:model_overview}
\end{figure}

Despite these advancements, text-to-SQL systems face several pressing challenges.

\textbf{Inefficiency when scaling to wide tables}. For schema linking, existing LLM-based methods typically input all columns from a table (or all tables), consuming substantial LLM window sizes and struggling to scale efficiently. Additionally, these approaches incur increasing costs with the growth in the number of columns and generally lack support for locating cell values, a critical feature given the frequent updates in real-world DBs.

\textbf{Design deficiencies in interaction-based methods}. Current methodologies based on execution refinement directly generate complete final SQL queries, which are coarse-grained and lack a step-by-step interpretable process for SQL generation. Furthermore, although multi-interaction-based methods hold promise, they exhibit significant shortcomings. For instance, \cite{Jiang-EMNLP-2023-StructGPT} does not provide tools for searching cell values, reducing its usability, \cite{Gu-Yu-2024-ArXiv-FUXI} encapsulates six SQL functions into separate functional tools, which adds unnecessary complexity when SQL execution alone would suffice. Additionally, these methods do not adequately address scalability concerns.

\textbf{Resource scarcity for annotating text-SQL data}. Current works emphasize prompt optimization by dynamically selecting exemplars based on similarity metrics. 
However, this approach assumes the availability of extensive training data as a candidate pool for exemplars while being unrealistic to expect user queries to consistently align with the training data distribution.
Moreover, these techniques require large annotated datasets that are costly to create, making it crucial to explore methods for low-resource settings.

Inspired by \cite{Xiong-Guanming-2024-ArXiv-Interactive-KBQA}, we propose the Interactive-\underline{T}ext-\underline{to}-\underline{S}QL framework, which leverages the reasoning capabilities of LLMs to interact with DBs through a step-by-step, conversational process.
As depicted in Figure \ref{fig:model_overview}, this framework conceptualizes the LLM as an agent and DBs as the environment, operating under a thought-action paradigm. Specifically, the LLM is required to think and then act, interacting with the DBs through a specially designed toolkit. We provided only two annotated exemplars with complete interactive processes as demonstrations for in-context learning, prompting the LLM to complete the task.
Comprehensive experiments on the Spider-Dev, BIRD-Dev, and their variant datasets showcased that our approach achieves significant results with minimal exemplar input. 

Our primary contributions are summarized as follows: 
\begin{itemize}[noitemsep] 
    \item Propose the Interactive-T2S, a novel framework for generating SQL queries through multi-turn interactions with DBs using LLMs.
    \item Design a unified interaction logic with four general tools that can effectively handle wide tables.
    \item Proof through extensive experiments that our method performs exceptionally well with just two exemplars.
\end{itemize}
\section{Related Work}

Recently, large language models (LLMs) have demonstrated remarkable reasoning capabilities \cite{Ouyang-Long-NeurIPS-2022-InstructGPT,OpenAI-2023-GPT4}, offering new opportunities for text-to-SQL systems. 
Current LLM-based methods can generally be categorized into two aspects: prompt optimization-based and interaction-based.

\textbf{Prompt optimization-based} approaches enhance SQL query generation by optimizing prompts for LLMs through schema linking and exemplar selection.
\textbf{Schema linking} aligns natural language questions with DB schema elements. Methods like \cite{Pourreza-Mohammadreza-NeurIPS-2023-DIN-SQL} input entire DB schema and questions into the model to identify relevant tables and columns. Hierarchical classification approaches by \cite{Lee-Dongjun-2024-ArXiv-MCS-SQL} first select relevant tables, then pinpoint columns. \cite{Tan-Zhao-LREC-COLING-2024-SL+CC+RS} show focusing on tables alone can outperform methods targeting columns. However, these models struggle with scalability due to LLMs' limited window sizes.
\textbf{Exemplar selection} involves choosing similar questions and queries to guide the model. Techniques by \cite{Guo-Chunxi-PRICAI-2023-DESEM} and \cite{Nan-Linyong-EMNLP-2023-SD+SA+Voting} utilize structural and syntactic similarities. \cite{Gao-Dawei-VLDB-2024-DAIL-SQL,Li-Zhishuai-2024-ArXiv-PET-SQL} prioritize candidates based on question and SQL similarities, while \cite{Zhang-Hanchong-EMNLP-2023-ACT-SQL} uses a hybrid approach. High-diversity demonstrations are explored by \cite{Wang-Dingzirui-2024-ArXiv-FUSED} to improve retrieval systems. However, these methods often assume complete training dataset access, limiting practical applicability.

\textbf{Interaction-based} approaches like \cite{Yao-Shunyu-ICLR-2023-ReAct} guide LLMs to interact with the environment to accomplish tasks.
Some works focus on refining SQL based on \textbf{execution results}. \cite{Shi-Freda-EMNLP-2022-MBR-Exec} introduced a framework that executes sampled SQL queries to select the most accurate translation based on execution risk. \cite{Ni-Ansong-ICML-2023-LEVER} incorporated a learned verifier estimating query correctness. \cite{Chen-Xinyun-ICLR-2024-SELF-DEBUGGING} developed a method for LLMs to autonomously debug and refine SQL by examining execution outcomes. \cite{Guo-Chunxi-ICONIP-2023-RAG+SP&DRC} applied a dynamic revision chain that uses execution feedback for correcting SQL semantic discrepancies.
Similar to our work, \cite{Jiang-EMNLP-2023-StructGPT} treats the \textbf{LLM as agent} and designs tools for extracting relevant information to interact with DBs. \cite{Gu-Yu-2024-ArXiv-FUXI} designs navigational and functional tools, but their tool design and interaction logic are suboptimal.

\section{Approach}

\begin{figure*}[htbp]
    \centering
    \includegraphics[width=0.87\textwidth,page=2]{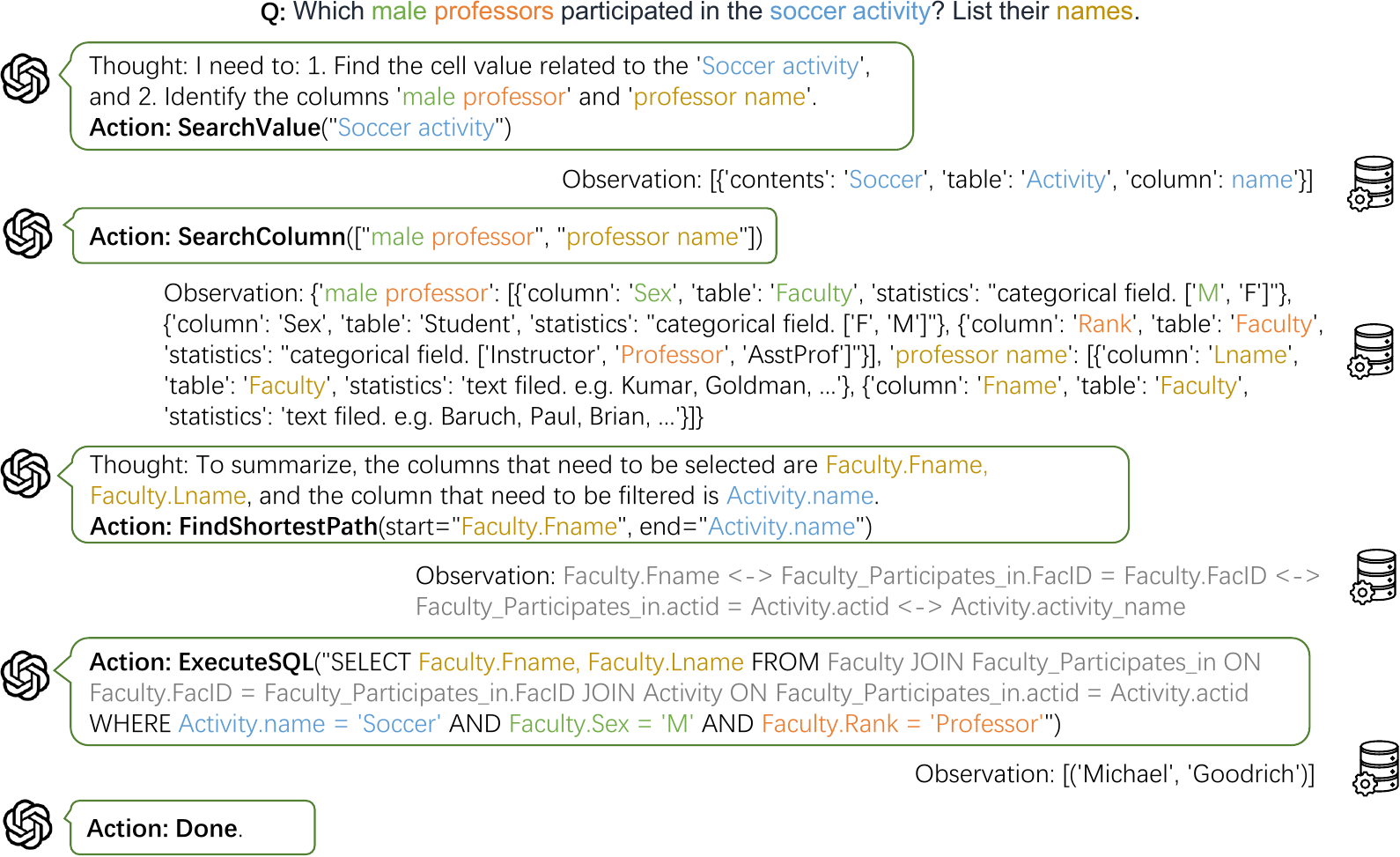}
    \caption{An example of the interactive process. \protect\footnotemark}
    \label{fig:dialog_overview}
\end{figure*}
\footnotetext{Due to limited space, some thought processes were omitted.}

\subsection{Problem Formulation}

This study investigates the text-to-SQL task.
A relational database (DB) is formally represented as $\mathcal{D} = \{\mathcal{T}, \mathcal{C}, \mathcal{V}\}$, where $\mathcal{T} = \{t_1, t_2, \ldots, t_{|\mathcal{T}|}\}$ is a set of tables, $\mathcal{C} = \{c_1, c_2, \ldots, c_{|\mathcal{C}|}\}$ is a set of columns, and $\mathcal{V} = \{v_1, v_2, \ldots, v_{|\mathcal{V}|}\}$ is a set of cell values. Each table $t_i$ comprises a set of columns $c_i$ and each column $c_i$ encompasses a set of cell values $v_i$.

Following \cite{Li-Haoyang-AAAI-2023-RESDSQL}, we further define foreign key relations $\mathcal{R} = \{(c^i_k, c^j_h) | c^i_k, c^j_h \in \mathcal{C}\}$, where each pair $(c^i_k, c^j_h)$ denotes a foreign key relation between column $c^i_k$ in table $i$ and column $c^j_h$ in table $j$. The database schema $\mathcal{S} = \{\mathcal{T}, \mathcal{C}, \mathcal{R}\}$ constitutes a set of tables, columns, and foreign key relations.

Formally, given a question $q$ and a database $\mathcal{D}$, the objective of the text-to-SQL task is to translate the question $q$ into a SQL query $l$ that can be executed on $\mathcal{D}$ to answer the question.

\subsection{Overview}

Recent advancements in large language models (LLMs) have highlighted their impressive capabilities in few-shot learning and logical reasoning. Nevertheless, designing scalable solutions for interpretable and step-by-step SQL generation in low-resource scenarios remains challenging.
In response, we introduce Interactive-T2S, a novel interactive method for text-to-SQL translation. This method treats the LLM as an agent interacting with a database environment, enhancing SQL generation through structured dialogic interactions. We developed a unified interaction logic with four generic tools to help LLMs identify relevant information and discern relationships across multiple tables. The example in Figure \ref{fig:dialog_overview} illustrates this interactive process. Different colors highlight how the corresponding elements can be located.

\subsection{Tools for Database}

We break down the process of generating SQL into three steps: searching for relevant columns and cell values, identifying the join relationships between tables where columns reside, and refining the prediction based on the execution.
In line with this principle, we introduce the following four tools.

\textbf{SearchColumn(semantic)} enhances the efficiency of LLMs by identifying the most relevant columns and excluding non-essential data. It concatenates and vectorizes the names and descriptions of each column, then ranks these columns according to their similarity to the parameter \texttt{semantic}.
Furthermore, following the methodology proposed by \cite{Li-Zhishuai-2024-ArXiv-PET-SQL}, we calculate and return the statistical characteristics of each column's cell values.

\textbf{SearchValue(value, table=None, column=None)} is designed to locate cell values across the entire DB.
Similar to the fuzzy match tool described in \cite{Gu-Yu-2024-ArXiv-FUXI}, we utilize BM25 to search for cell values within the DB. If the parameters \texttt{table} or \texttt{column} are specified, the tool will conduct searches within the designated table or column.

\textbf{FindShortestPath(start, end)} is designed to efficiently identify the shortest path between two columns within a DB schema, based on foreign key relationships. 
Existing methods heavily rely on the intrinsic capabilities of LLM to perform joins across multiple tables, which becomes impractical with scenarios involving extensive joins or a large number of columns.
In contrast, the number of tables requiring joins is dictated solely by the DB schema design, rather than the semantic content of the question. 
By modeling the DB schema as an undirected graph where columns are nodes and edges are defined by column relationships and foreign keys, this tool simplifies multi-table joins and reduces LLM workload.

\textbf{ExecuteSQL(sql)} provides the capability to execute SQL queries directly, offering significant flexibility.

\subsection{Interactive Process}

Given a question $q$, we first construct a prompt text:
\begin{equation}
    \text{Prompt} = \{\text{Inst}, E, S_q, q\}
\end{equation}

where $\text{Inst}$ denotes the pre-written instruction text, which encompasses descriptions of tools, usage guidelines, and the required format. $E=[(S_0, e_0, ...e_n), ...]$ represents a list of demonstrations, each consisting of a database schema $S_i$ and $n$ exemplars $e$ with full interactive process.

In each turn $T$, we prompt the LLM generate an action based on the $\text{Prompt}$ and the historical interaction $H$. Specifically, this process is described by:
\begin{equation}
    \label{eq:interactive_process}
    a_T = \text{LLM}(\{\text{Prompt}, H\})
\end{equation}
\begin{equation}
    H = \{c_0,a_0,o_0, ...,c_{T-1}, a_{T-1}, o_{T-1}\}
\end{equation}

where $c$ denotes the intermediate thought process, an action $a$ belongs to the set \{SearchColumn, SearchValue, FindShortestPath, ExecuteSQL, Done\}, and the observation $o$ results from executing an action, defined as $o_T = \text{Tool}(a_T)$.

\subsection{General Solution for Text-to-SQL}

We propose a general and unified interaction logic for generating SQL queries, as the example illustrated in Figure \ref{fig:dialog_overview}.
The process begins with \textbf{locating elements}. Initially, the LLM is tasked with decomposing the user's question into a conceptual plan ($c_0$), which is flexibly designed to adapt to the semantics of the question, enhancing comprehension for both LLMs and humans. Following this, the agent is required to generate a thought $c$ and an action $a_T$ aimed at identifying pertinent columns and cell values within the DB.
The next phase is \textbf{joining tables}. As the question shown in Figure~\ref{fig:dialog_overview}, the selected columns are \texttt{Faculty.Fname} and \texttt{Faculty.Lname}, while the filtered column is \texttt{Activity.name} (with the constraint ``activity = soccer''). Note that for the tool, \texttt{start} and \texttt{end} are interchangeable since the schema is treated as an undirected graph; the distinction between selected and filtered columns is only for organizing the thought process.
The final phase is \textbf{execute SQL}, where the LLM executes the constructed SQL query to retrieve the desired results. This query execution is deemed the final output.

\section{Experiment}

\subsection{Dataset}

\begin{table}[ht]
    \centering
    \caption{Statistics of the datasets.}
    \label{tab:data_statistics}
    \scalebox{0.95}{

    \begin{tabular}{lcccccc}
        \toprule
        \multicolumn{1}{c}{\multirow{2}{*}{\textbf{Dataset}}} & \multicolumn{2}{c}{\textbf{\#Item}} & \multicolumn{2}{c}{\textbf{\#DB}} & \multicolumn{2}{c}{\textbf{Statistics}} \\ \cline{2-7} 
        \multicolumn{1}{c}{} & \textbf{Train} & \textbf{Dev} & \textbf{Train} & \textbf{Dev} & \textbf{CVR} & \textbf{CVCR} \\ \midrule
        Spider & 8,659 & 1,034 & 146 & 20 & 10.25 & 87.14 \\
        \quad - DK & - & 535 & - & 10 & 11.59 & 40.79 \\
        \quad - Syn & 7,000 & 1,034 & 140 & 20 & 10.44 & 76.06 \\
        \quad - Realistic & - & 508 & - & 19 & 16.54 & 84.96 \\ \midrule
        BIRD & 9,428 & 1,534 & 69 & 11 & 68.38 & 63.68 \\
        FinC (Original) & \multirow{3}{*}{-} & \multirow{3}{*}{106} & \multirow{3}{*}{-} & \multirow{3}{*}{1} & 68.87 & 33.66 \\
        \quad - (SQLC) &  &  &  &  & 70.75 & 32.04 \\
        \quad - (DataC) &  &  &  &  & 70.75 & 34.95 \\ Mini-Dev & - & 500 & - & 11 & 70.60 & 59.63 \\ \bottomrule
    \end{tabular}

    }
\end{table}

\textbf{Spider} \cite{Yu-Tao-EMNLP-2018-Spider} is a widely used cross-domain text-to-SQL benchmark. We utilize the development set. Building on this, \textbf{Spider-DK} \cite{Gan-Yujian-EMNLP-2021-Spider-DK} challenges parsers with domain knowledge reasoning and real-world question paraphrases to evaluate cross-domain generalization. \textbf{Spider-Syn} \cite{Gan-Yujian-ACL-2021-Spider-SYN} introduces synonyms for schema tokens to prevent reliance on string matching. \textbf{Spider-Realistic} \cite{Den-Xiang-NAACL-2021-Spider-Realistic} further challenges models by revising questions to omit explicit column names, thus testing text-table alignment capabilities.

\textbf{BIRD} \cite{Li-Jinyang-NeurIPS-2024-BIRD} is notable for its complexity, featuring intricate instructions over highly complex databases. BIRD has two settings: with and without external knowledge evidence (oracle knowledge), which provides specific information needed for answering questions. We report results on both the development and test sets in our experiments.
\textbf{BIRD-Financial Corrected (FinC)} \cite{Niclas-Wretblad-ACL-2024-short-Effects-of-Noise} addresses the issue of noise in the BIRD benchmark, particularly the uneven distribution of incorrect SQL queries that affects its reliability. They have revised the dataset under the financial domain, which includes 106 question and SQL query pairs, representing approximately 7.5\% of the development data. The correction includes SQL-only corrections \textbf{(SQLC)} and corrections for both SQL queries and noisy questions \textbf{(DataC)}.
\textbf{BIRD Mini-Dev} dataset offers 500 high-quality text-SQL pairs derived from community feedback. We utilize the SQLite version.
Following \cite{Gu-Yu-2024-ArXiv-FUXI}, we argue that using oracle knowledge is unreasonable; however, we still report the results of both for comparison.

The detailed statistics are listed in Table \ref{tab:data_statistics}.
We also provide statistics for the latest Spider 2.0 dataset in Appendix \ref{app_sec:spider2_results}.

\subsection{Baselines}

To comprehensively evaluate our approach, we have selected various state-of-the-art (SOTA) baseline models.

\textbf{Fine-tuning (FT) on full data}. CodeS \cite{Li-Haoyang-SIGMOD-2024-CodeS}, a series of pre-trained language models, addresses schema linking and domain adaptation through incremental pre-training on a SQL-centric corpus, strategic prompts, and bi-directional data augmentation. We select the best results in the Supervised Fine-Tuning (SFT) setting.

\textbf{Prompt-based methods}. We selected recent prompt-based methods using GPT-4 models. In the context of in-context learning \cite{Dong-Qingxiu-2024-ArXiv-Survey-In-context-Learning}, these can be categorized into \textbf{Selection-based} and \textbf{Fixed Few-shot}: the former dynamically selects exemplars from training data based on similarity, while the latter uses a fixed set of examples.

Notably, few works have addressed the BIRD dataset series without oracle knowledge. Therefore, we reimplemented several baselines following \cite{Niclas-Wretblad-ACL-2024-short-Effects-of-Noise}.

\subsection{Implementation Details}
\label{sec:imp_details}

We employ OpenAI's gpt-4o-2024-05-13 as the LLM agent in our experiments. To prevent excessive token consumption, we set a maximum of 12 interaction rounds and restrict the length of each observation's returned results.
The few-shot exemplars are manually constructed. Specifically, we randomly selected and analyzed several challenging cases, annotating two representative examples each for the Spider and BIRD dataset series.
Details regarding the design of prompt texts and the implementation of tools can be found in Appendix~\ref{app_sec:more_implementation_details} and Appendix~\ref{app_sec:tools_implementation_details}.

\subsection{Evaluation Metrics}

Following \cite{Qu-Ge-2024-ArXiv-TA-SQL} and \cite{Shen-Zhili-2024-ArXiv-AST-based-Ranking}, we report exact match accuracy (EM) \cite{Yu-Tao-EMNLP-2018-Spider} and execution accuracy (EX) \cite{Zhong-Ruiqi-EMNLP-2020-Semantic-Evaluation} for the Spider dataset series. EM requires each component of the predicted SQL to match the gold SQL, but it cannot handle cases where multiple correct answers exist \cite{Sun-Ruoxi-2024-ArXiv-SQL-PaLM,Zhang-Hanchong-EMNLP-2023-ACT-SQL}, so we report it only for reference. EX, which requires the execution result of the predicted SQL to be correct, is generally more precise. For the BIRD-Dev and FinC datasets, we use the EX metric. For the Mini-Dev dataset, we also report the Soft F1-Score \cite{Li-Jinyang-NeurIPS-2024-BIRD}.\footnote{We omit the valid efficiency score (VES) and the Reward-based Valid Efficiency Score (R-VES) as these metrics depend on hardware performance.}

\subsection{Main Results}

\begin{table*}[htpb]
    \centering
    \scalebox{1}{

    \begin{tabular}{clcccccccc}
        \toprule
        \multicolumn{2}{c}{\multirow{2}{*}{\textbf{Method}}} & \multicolumn{2}{c}{\textbf{Spider-Dev}} & \multicolumn{2}{c}{\textbf{Spider-DK}} & \multicolumn{2}{c}{\textbf{Spider-Syn}} & \multicolumn{2}{c}{\textbf{Spider-Realistic}} \\ \cline{3-10} 
        \multicolumn{2}{c}{} & \textbf{EM} & \textbf{EX} & \textbf{EM} & \textbf{EX} & \textbf{EM} & \textbf{EX} & \textbf{EM} & \textbf{EX} \\ \midrule
        FT SOTA & CodeS (SFT) \cite{Li-Haoyang-SIGMOD-2024-CodeS} & - & 85.4 & - & 70.7 & - & 77.0 & - & 83.1 \\ \midrule
        \multirow{3}{*}{Selection-based} & DAIL-SQL \cite{Gao-Dawei-VLDB-2024-DAIL-SQL} & 71.9 & 83.6 & - & - & - & - & - & 76.0 \\
         & AST Norm\tablefootnote{For Spider-Dev, we reprint GPT-4+Graphix-T5. For others, we chose GPT-4+FastRAT$_{ext}$.} \cite{Shen-Zhili-2024-ArXiv-AST-based-Ranking} & 77.3 & 86.6 & 59.1 & 72.3 & 61.3 & 74.4 & 66.1 & 80.9 \\
         & PURPLE \cite{Ren-Tonghui-2024-ArXiv-PURPLE} & 80.5 & \textbf{87.8} & 61.7 & 75.3 & 63.3 & 74.0 & 71.1 & 79.9 \\ \midrule
        \multirow{3}{*}{Fixed Few-shot} & TA-SQL (7-shot)\tablefootnote{The task-aligned logical synthesis module uses 7-shot.} \cite{Qu-Ge-2024-ArXiv-TA-SQL} & 44.3 & 85.0 & - & 72.9 & - & - & - & 79.5 \\
         & SL+CC+RS (6-shot) \cite{Tan-Zhao-LREC-COLING-2024-SL+CC+RS} & - & 86.2 & - & 67.2 & - & 78.1 & - & \textbf{82.8} \\
         & \textbf{Ours (2-shot)} & 52.0 & 84.8 & 41.5 & \textbf{75.5} & 44.9 & \textbf{78.7} & 55.7 & 81.3 \\ \bottomrule
    \end{tabular}

    }
    \caption{Results on Spider-Dev and its variants.}
    \label{tab:main_res_Spider_series}
\end{table*}

\begin{table}[ht]
    \centering
    \caption{Results on BIRD-Dev.\tablefootnote{Results tagged with $\sharp$ are reprinted from the official leaderboards: \url{https://BIRD-bench.github.io} and \url{https://github.com/BIRD-bench/mini\_dev}.}}
    \label{tab:main_res_BIRD_dev}
    \scalebox{1}{

    \begin{tabular}{clc}
        \toprule
        \multicolumn{2}{c}{\textbf{Model}}                         & \textbf{EX}    \\ \midrule
        
        \multicolumn{3}{c}{\textit{w/ Oracle Knowledge}}                    \\ \midrule
        FT SOTA                          & CodeS (SFT)             & 58.47          \\
        SOTA                             & Distillery + GPT-4o $\sharp$     & \textbf{67.21} \\ \midrule
        \multirow{2}{*}{Selection-based} & DAIL-SQL + GPT-4        & 54.76          \\
                                         & SuperSQL \cite{Li-Boyan-2024-ArXiv-NL2SQL360-SuperSQL}                & 58.50          \\ \midrule
        \multirow{3}{*}{Fixed Few-shot}        & TA-SQL + GPT-4          & 56.19          \\
                                         & DIN-SQL + GPT-4 \cite{Pourreza-Mohammadreza-NeurIPS-2023-DIN-SQL} & 50.72          \\
                                         & \textbf{Ours}           & 60.76          \\ \midrule
        \multicolumn{3}{c}{\textit{w/o Oracle Knowledge}}                  \\ \midrule
        FT SOTA                          & CodeS (SFT)             & 47.91          \\
        SOTA                             & ExSL + granite-20b-code $\sharp$ & 51.69          \\ \midrule
        \multirow{3}{*}{Fixed Few-shot}        & StructGPT \cite{Jiang-EMNLP-2023-StructGPT}               & 31.80          \\
                                         & FUXI \cite{Gu-Yu-2024-ArXiv-FUXI}                   & 42.90          \\
                                         & \textbf{Ours}           & \textbf{54.56} \\ \bottomrule
    \end{tabular}

    }
\end{table}

\begin{table}[ht]
    \centering
    \caption{Results on BIRD-FinC. The EX metric is reported.\tablefootnote{Results tagged with $\dagger$ indicates we reimplemented the results.}}
    \label{tab:main_res_BIRD_finc}
    \scalebox{1}{
    
    \begin{tabular}{lccc}
        \toprule
        \textbf{Model} & \textbf{Original} & \textbf{SQLC} & \textbf{DataC} \\ \midrule
        \multicolumn{4}{c}{\textit{w/ Oracle Knowledge}}                \\ \midrule
        Zero-shot (GPT-4)    & 38.09          & 48.11          & 55.66          \\
        Zero-shot (GPT-4o) $\dagger$   & 50.00          & 57.55          & 62.26 \\
        DIN-SQL (GPT-3.5)    & 34.91          & 38.68          & 47.16          \\
        DIN-SQL (GPT-4o) $\dagger$     & 43.40          & 50.94          & 65.09 \\
        \textbf{Ours}        & \textbf{54.72} & \textbf{65.09} & \textbf{69.81} \\ \midrule
        \multicolumn{4}{c}{\textit{w/o Oracle Knowledge}}              \\ \midrule
        Zero-shot (GPT-4o) $\dagger$   & 26.42          & 31.13          & 35.85 \\
        DIN-SQL (GPT-4o) $\dagger$     & 39.62          & 47.17          & 56.60 \\
        \textbf{Ours}        & \textbf{44.34} & \textbf{49.06} & \textbf{58.49} \\ \bottomrule
    \end{tabular}

    }
\end{table}

\begin{table}[ht]
    \centering
    \caption{Results on BIRD Mini-Dev.}
    \label{tab:main_res_BIRD_minidev}
    \scalebox{1}{

    \begin{tabular}{lcc}
        \toprule
        \multicolumn{1}{c}{\textbf{Model}} & \textbf{EX}    & \textbf{Soft F1-Score} \\ \midrule
        \multicolumn{3}{c}{\textit{w/ Oracle Knowledge}}                    \\ \midrule
        GPT-4 $\sharp$                             & 47.80          & 52.69                  \\
        TA + GPT-4-turbo $\sharp$                  & 58.00          & 62.40                  \\
        TA + GPT-4o $\sharp$                       & \textbf{63.00} & \textbf{66.97}         \\
        \textbf{Ours}                      & 58.80          & 63.07                  \\ \midrule
        \multicolumn{3}{c}{\textit{w/o Oracle Knowledge}}                   \\ \midrule
        Zero-shot (GPT-4o) $\dagger$       & 28.00          & 31.99                  \\
        DIN-SQL (GPT-4o) $\dagger$         & 36.80          & 40.60                  \\
        \textbf{Ours}                      & \textbf{46.60} & \textbf{50.75}        \\ \bottomrule
    \end{tabular}

    }
\end{table}

\paragraph{Analysis of Spider-Dev and its variants}

\begin{table}[ht]
    \centering
    \caption{Statistics of the average number of tables (Tb) per DB, columns (Col) per table, foreign keys (FK) per DB, and schema tokens (STk) per DB.\tablefootnote{This paper uses the OpenAI GPT-4o tokenizer by default.}}
    \label{tab:dataset_schema_statistics}
    \scalebox{1}{

    \begin{tabular}{lcccc}
        \toprule
        \textbf{Dataset (Dev)} & \textbf{Tb/DB} & \textbf{Col/Tb} & \textbf{FK/DB} & \textbf{STk/DB} \\ \midrule
        Spider       & 4.00          & 5.49           & 3.25          & 307                \\
        BIRD         & 6.82          & 10.64          & 9.55          & 3,324               \\ \bottomrule
    \end{tabular}

    }
\end{table}

Experimental results are presented in Table \ref{tab:main_res_Spider_series}. Our method shows competitive performance across the Spider-DK, Syn, and Realistic datasets. Selection-based methods, which choose similar exemplars from training data, outperform fixed few-shot approaches, as confirmed by the EM metric. However, the performance gap narrows significantly across the three variant datasets, suggesting selection-based methods assume similar data distributions between development and training sets, indicating their inadequate generalization capabilities.

In fixed few-shot methods, both TA-SQL and SL+CC+RS perform well because they utilize the entire DB schema (including all columns) as prompt text for schema linking, which is feasible due to the relatively small size of the DB. 
Table \ref{tab:dataset_schema_statistics} highlights that Spider-Dev has only an average of 307 schema tokens per DB.
Despite the complexity of TA-SQL, which uses intricate modules for generating SQL and pandas-like APIs for reasoning, our design employs a simple yet effective unified interaction logic. SL+CC+RS performs worse when incorporating domain knowledge (DK dataset), highlighting our approach's superior generalization.

We contend that our method's performance on Spider-Dev is limited by the dataset's simplicity and ambiguity, as elaborated in subsequent sections on difficulty analysis and error analysis, respectively.
We also conducted experiments on the Spider 2.0 dataset, as detailed in Appendix \ref{app_sec:spider2_results}.

\paragraph{Analysis of BIRD-Dev and its variants}

In our analysis, following the critiques by \cite{Gu-Yu-2024-ArXiv-FUXI} regarding the impracticality of oracle knowledge in real-world applications, we focus primarily on settings without oracle knowledge.
Our method surpasses the SOTA on the BIRD-Dev dataset by 2.87\%, as shown in Table \ref{tab:main_res_BIRD_dev}.\footnote{Our leaderboard score is 54.11.}
Furthermore, we reimplemented the Zero-shot and DIN-SQL on both Mini-Dev and BIRD-FinC dataset, as conducted by \cite{Niclas-Wretblad-ACL-2024-short-Effects-of-Noise}. The results are provided in Table \ref{tab:main_res_BIRD_minidev}. 
Across all datasets, our method sets a new standard in the setting without oracle knowledge, establishing new SOTA results.

\subsection{The Difficulty Analysis of Locating Cell Values}

The difficulty of SQL generation is often superficially measured by the presence of specific SQL keywords, as in prior work~\cite{Yu-Tao-EMNLP-2018-Spider,Li-Jinyang-NeurIPS-2024-BIRD}. In contrast, we propose a more rigorous and fine-grained assessment based on the challenges of accurately locating schema elements, identifying cell values, and performing table joins. In this section, we focus on the problem of cell value localization, while the other aspects are discussed in subsequent sections.

Inspired by the Mention Cover Rate metric from knowledge-based QA \cite{Xiong-Guanming-2024-ArXiv-Interactive-KBQA}, we introduce two metrics: Cell Value Rate (CVR) and Cell Value Cover Rate (CVCR). CVR measures the proportion of SQL queries with value constraints, while CVCR indicates how often these constraint values appear directly in the questions. These metrics help quantify the difficulty of locating cell values.

Statistics presented in Table \ref{tab:data_statistics} reveal significant disparities between the Spider and BIRD dataset series. 
Specifically, the BIRD series requires the identification of cell values in approximately seven times as many cases as the Spider series.
Notably, in Spider-Dev, around 87\% of cases requiring cell value identification include the golden cell value directly in the question, simplifying the SQL generation process. 
Conversely, Spider-DK reduces the CVCR to 40.79\%, substantially increasing the complexity and leading to a notable performance drop in SL+CC+RS. 
These findings underscore the necessity of developing tools like the SearchValue tool in our framework, which aids LLMs in pinpointing cell values.

\subsection{The Efficiency Analysis of Schema Linking}

\begin{table}[ht!]
    \centering
    \caption{Statistics of the prompt token consumption in DIN-SQL, which consists of four modules: Schema Linking (SLink), Classification \& Decomposition (QClDe), SQL Generation (SQLGen), and Self-correction (SelfC).}
    \label{tab:tokens_din_sql}
    \scalebox{1}{

    \begin{tabular}{cccc}
        \toprule
        \multicolumn{4}{c}{\textbf{Tokens of Fixed Prompt Text}}        \\ \midrule
        \textbf{SLink} & \textbf{QClDe} & \textbf{SQLGen} & \textbf{SelfC} \\ \midrule
        3,411           & 4,028          & 3,170            & 997            \\ \midrule \midrule
        \multicolumn{2}{c}{\multirow{2}{*}{\textbf{Total (Fixed)}}} & \multicolumn{2}{c}{\textbf{Total}} \\
        \multicolumn{2}{c}{}           & \textbf{Spider-Dev} & \textbf{BIRD-Dev}  \\ \midrule
        \multicolumn{2}{c}{11,606}      & 12,834           & 21,622          \\ \bottomrule
    \end{tabular}

    }
\end{table}

\begin{table}[ht!]
    \centering
    \caption{Statistics of token consumption of fixed prompt text (\#Tk of Fixed), including tool descriptions (Desc) and exemplars (Exem), as well as the average number of interaction rounds (Avg. R), average total number of tokens (Avg. Tk), and average cost (Avg. \$). The term ``kg'' refers to oracle knowledge.}
    \label{tab:tokens_ours_avg_round_cost}
    \scalebox{1}{

    \begin{tabular}{lccccc}
        \toprule
        \multicolumn{1}{c}{\multirow{2}{*}{\textbf{\begin{tabular}[c]{@{}c@{}}Dataset\\(Dev)\end{tabular}}}} &
          \multicolumn{2}{c}{\textbf{\#Tk of Fixed}} &
          \multirow{2}{*}{\textbf{Avg. R}} &
          \multirow{2}{*}{\textbf{Avg. Tk}} &
          \multirow{2}{*}{\textbf{Avg. \$}} \\ \cline{2-3}
        \multicolumn{1}{c}{} & \textbf{Desc}    & \textbf{Exem} &      &      &      \\ \midrule
        Spider           & \multirow{3}{*}{1,160} & 2,308              & 4.36 & 4,634 & 0.10 \\
        BIRD (w/ kg)     &                       & 1,608              & 5.59 & 4,474 & 0.12 \\
        BIRD (w/o kg)    &                       & 1,690              & 5.75 & 4,715 & 0.13 \\ \bottomrule
    \end{tabular}

    }
\end{table}

In this section, we analyze the efficiency of schema linking using DIN-SQL as a case study. 
Table \ref{tab:tokens_din_sql} lists the number of fixed prompt text (input) tokens across the four modules of DIN-SQL, as well as the estimated total prompt tokens per case based on the average tokens per DB detailed in Table \ref{tab:dataset_schema_statistics}. Given that DIN-SQL utilizes the complete DB schema in each module, the average prompt tokens per case are calculated as the sum of total fixed tokens and four times the average tokens per DB, resulting in about 12.8k and 21.6k for Spider-Dev and BIRD-Dev, respectively.

Comparatively, our method, as presented in Table \ref{tab:tokens_ours_avg_round_cost}, requires only 4.6k and 4.7k tokens per case, which corresponds to approximately 36\% and 22\% of the tokens required by DIN-SQL, respectively.
This efficiency stems from our method's dynamic retrieval of necessary information, which is not affected by the length of the DB schema, demonstrating our method's scalability.
It is important to note that decoder-only LLMs employ causal decoding with KV-Cache techniques \cite{Shi-Luohe-2024-ArXiv-Review-KV-Cache}, which enables computational efficiency when processing repetitive prefixes. Our interactive approach leveraging the OpenAI API benefits from these optimizations in terms of reduced computational costs.

\subsection{Ablation Study}

\begin{table}[ht]
    \centering
    \caption{Distribution of the number of tables involved in golden SQL queries.}
    \label{tab:join_statistics}
    \scalebox{1}{

    \begin{tabular}{lccccc}
        \toprule
        \multicolumn{1}{c}{\multirow{2}{*}{\textbf{\begin{tabular}[c]{@{}c@{}}Dataset\\(Dev)\end{tabular}}}} & \multicolumn{4}{c}{\textbf{Number of Tables}} & \multirow{2}{*}{\textbf{Avg.}} \\ \cline{2-5}
        \multicolumn{1}{c}{} & \textbf{1} & \textbf{2} & \textbf{3} & \textbf{4+} &      \\ \midrule
        Spider       & 60.54      & 30.95      & 6.96       & 1.55        & 1.50 \\
        BIRD             & 25.68      & 58.93      & 13.95      & 1.43        & 1.92 \\ \bottomrule
    \end{tabular}

    }
\end{table}

\begin{table}[ht!]
    \centering
    \caption{The impact of the FindShortestPath tool. The EX metric is reported.}
    \label{tab:findpath_ablation}
    \scalebox{1}{

    \begin{tabular}{lcccc}
        \toprule
        \multicolumn{1}{c}{\multirow{2}{*}{\textbf{Dataset}}} & \multicolumn{3}{c}{\textbf{Number of Tables}} & \multirow{2}{*}{\textbf{Avg.}} \\ \cline{2-4}
        \multicolumn{1}{c}{} & \textbf{2} & \textbf{3} & \textbf{4+} &       \\ \midrule
        Spider           & 80         & 84         & 76          & 80.00 \\
        \quad w/o findpath         & 74         & 80         & 54          & 69.33 \\
        Gain                 & 6          & 4          & 22          & 10.67 \\
        BIRD             & 70         & 58         & 40          & 58.00 \\
        \quad w/o findpath         & 62         & 57         & 28          & 51.33 \\
        Gain                 & 8          & 1          & 12          & 6.67  \\ \bottomrule
    \end{tabular}

    }
\end{table}

In this section, we explore the impact of the FindShortestPath tool. We categorized and sampled cases from Spider-Dev and BIRD-Dev based on the number of table joins (2, 3, 4+), creating two subsets, each containing 50 cases selected randomly or supplemented from variant datasets if necessary. 
As depicted in Table \ref{tab:findpath_ablation}, the tool significantly enhanced performance in scenarios requiring joins across four or more tables. 
Additionally, we analyzed the distribution of join table counts within the golden SQL queries, as listed in Table \ref{tab:join_statistics}. Notably, cases with four or more table joins constitute less than 2\% of both datasets, underscoring the substantial potential of our tool. 
Importantly, the complexity of a question's semantics does not necessarily correlate with the number of table joins, which depend solely on the DB's design. Hence, the FindShortestPath tool is designed to decouple the path-finding process from direct inference by LLMs from the DB schema, thereby alleviating unnecessary reasoning burdens and ensuring the system's scalability.
We also conducted the experimental results of open-source LLMs in Appendix \ref{app_sec:open_llm}.

\subsection{Error Analysis}

\begin{table}[!ht]
    \centering
    \caption{Distribution of error types.}
    \label{tab:error_analysis}
    \scalebox{1}{

    \begin{tabular}{lcc}
        \toprule
        \textbf{Name} & \textbf{Spider-Dev} & \textbf{BIRD-Dev}  \\ \midrule
        \multicolumn{3}{c}{Mismatch}                            \\ \midrule
        Golden Wrong          & 10              & 2             \\
        Golden Empty          & 35              & 4             \\
        Ambiguous             & 25              & 14            \\
        Distinct              & 6               & 8             \\
        Format                & 5               & 18            \\
        Lacking Info          & -               & 20            \\
        Total                 & 82              & 86            \\ \midrule
        \multicolumn{3}{c}{Error}                               \\ \midrule
        Selected Column       & 6               & 20            \\
        Filtered Column       & 3               & 4             \\
        Reasoning             & 9               & 10            \\
        Total                 & 18              & 34            \\ \bottomrule
    \end{tabular}

    }
\end{table}

In this section, we performed an error analysis by randomly sampling 100 cases from both the Spider-Dev and BIRD-Dev datasets, consistent with the original data distribution. The results are documented in Table \ref{tab:error_analysis}. 
We categorized the errors into two main types: Mismatch and Error. A Mismatch refers to cases where the generated SQL does not match the golden SQL but maintains semantic consistency with the question. Conversely, an Error indicates an incorrect SQL generation.

For Mismatch types:
\textbf{Golden Wrong} highlights mismatches where the golden SQL does not align with the question requirements. 
\textbf{Golden Empty} refers to scenarios in which the golden SQL query produces no results. Within our framework, this occurrence might lead the model to mistakenly perceive its generated query as incorrect during the intermediate steps. Such a misperception can lead to unnecessary interactions, ultimately resulting in erroneous outputs.
\textbf{Ambiguous} represents cases where multiple valid SQL queries could correctly answer the question, but differ from the single golden reference, highlighting the limitation of having only one reference solution.
\textbf{Distinct} denotes that the predicted and the golden differ by only one DISTINCT keyword, which is not explicitly required in the question. 
\textbf{Format} indicates correct queries that are hindered by format inconsistencies, such as NULL handling or time fields, resulting in different execution outcomes.
\textbf{Lacking Info} applies exclusively to the BIRD dataset, signifying cases where missing oracle knowledge makes the question challenging to answer accurately.

For Error types:
\textbf{Selected Column} points to inaccuracies in the prediction of the selected column, often due to numerous similar column names.
\textbf{Filtered Column} refers to errors in predicting the correct filtered column.
\textbf{Reasoning} indicates fundamental issues with the SQL structure, beyond mere column inaccuracies.

Surprisingly, only 18\% and 34\% of all erroneous cases are caused by reasoning errors. Therefore, we argue that different datasets possess their unique annotation styles, and it is exceedingly challenging to identify and align with these styles using very few demonstrations.
\begin{acks}
This work was supported by the National Key Research and Development Program of China under Grant No. 2023YFC3304404.
\end{acks}

\section{Conclusion}

Interactive-T2S is a text-to-SQL approach that leverages a LLM as an agent to generate SQL queries through multi-round interactions with a database. We designed a unified tool and interaction methodology for schema linking, cell value localization, table joining, and query refinement based on execution results. Additionally, we employed a few-shot learning strategy to guide the LLM in incrementally generating SQL queries. Experimental results demonstrate that our method achieves SOTA results with just two exemplars.
\section{Appendix}

This appendix provides detailed experimental results and offers further discussion.

\subsection{More Implementation Details}\label{app_sec:more_implementation_details}

\begin{figure*}[htbp]
    \centering
    \includegraphics[width=1\textwidth,page=1]{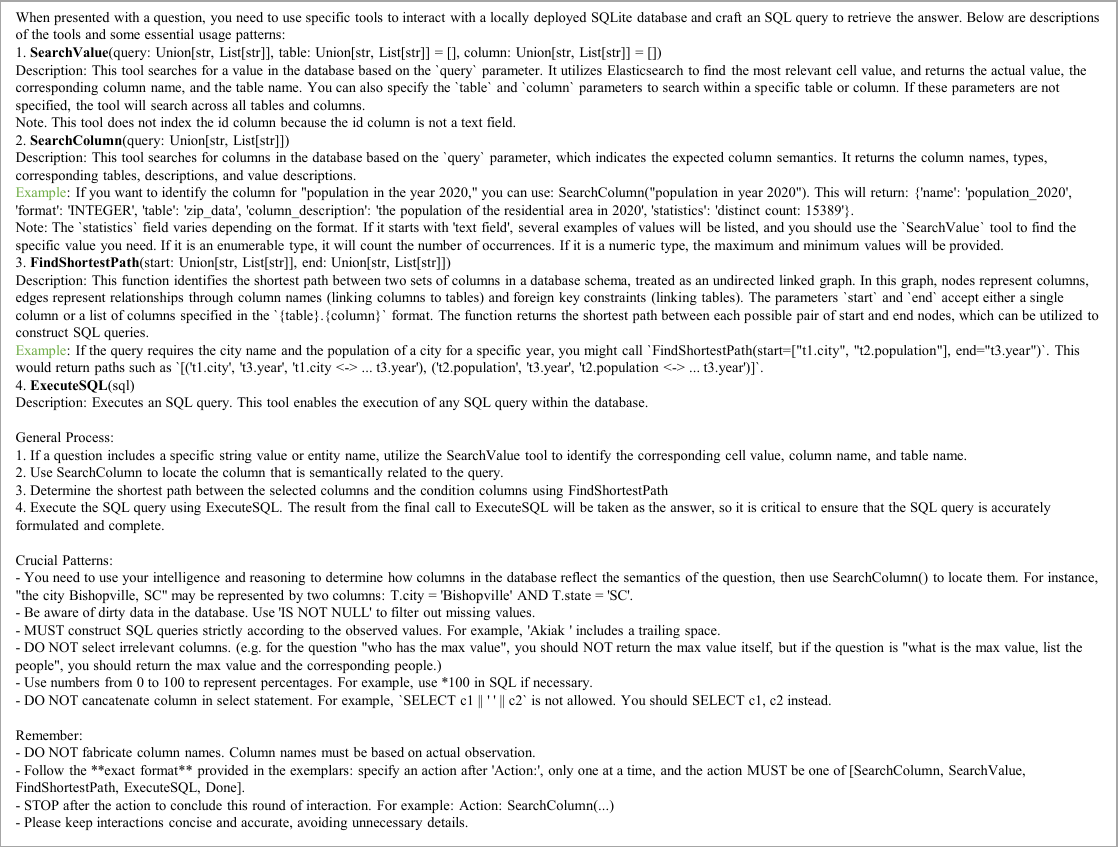}
    \caption{Prompt text.}
    \label{fig:prompt_text}
\end{figure*}

The prompt text is shown in Figure \ref{fig:prompt_text}. The few-shot exemplars are manually constructed. For Spider, we observed that queries related to the ``academic'' database returned empty results, which could mislead the LLM into thinking it made an error and continue exploring unnecessarily. Therefore, we selected one such case as an example.

\subsection{Tool Implementation Details}\label{app_sec:tools_implementation_details}

\textbf{SearchColumn(semantic)}. The tool SearchColumn is designed to rank database columns based on their relevance to the \texttt{semantic} parameter. 
For the Spider dataset, column and table names are embedded using the template ``a column named \{column\_name\} in table \{table\_name\}''. 
For the BIRD dataset, the template ``a column named \{column\_name\} in table \{table\_name\} about \{desc\}'' is used, where \{desc\} includes descriptions of the column provided by the dataset \cite{Li-Jinyang-NeurIPS-2024-BIRD}. 

We utilize the OpenAI \texttt{text-embedding-3-large} API to generate vectors and employ Chroma\footnote{\url{https://github.com/chroma-core/chroma}} to index and search.

The tool will return the following features for each column:
\begin{itemize}[noitemsep]
    \item \texttt{column\_name}: the name of the column.
    \item \texttt{table\_name}: the name of the table.
    \item \texttt{column\_type}: the data type of the column.
    \item \texttt{column\_desc}: the description of the column.
    \item \texttt{column\_statistics}: the statistics of the cell values in the column.
\end{itemize}

In our approach, the column\_name, table\_name, and column\_type are extracted through SQL queries. 
For describing the columns semantically, we adopt the ``semantic name'' as the ``column\_desc'' in the context of the Spider dataset, following the methodology outlined by \cite{Li-Haoyang-AAAI-2023-RESDSQL}. 
For the BIRD dataset, we utilize the column descriptions as provided within the original dataset. 
Furthermore, we enhance the representation of each column by computing statistical features from the cell values, an extension to the ``cell value reference'' presented by \cite{Li-Haoyang-AAAI-2023-RESDSQL}. 
Specifically, for text-based columns, we randomly sample cell values and return the first 100 characters; for numeric or date types, we calculate and return the maximum and minimum values. This enriched feature set aids in a deeper understanding and processing of column data.

\textbf{SearchValue(value)}. This tool is designed for searching the values within a column utilizing Elasticsearch, where only the text fields are indexed.

\textbf{FindShortestPath(start, end)}. This tool computes the shortest path between two nodes in a graph. It leverages the NetworkX \cite{Hagberg-2008-NetworkX} library, a powerful tool for the analysis of complex networks.

\textbf{ExecuteSQL(sql)}. This tool executes a provided SQL query using the SQLite3 library in Python 3.

\subsection{System Configurations}

\begin{table}[ht]
    \centering
    \caption{Assignments of hyper-parameters for inference.}
    \label{app_tab:hyper_infer}
    \begin{tabular}{lc}
        \toprule
        \textbf{Parameter}   & \textbf{Value} \\ \midrule
        model & gpt-4o-2024-05-13 \\
        temperature & 0.7   \\
        top\_p      & 0.95  \\
        n           & 1     \\
        stop  & [``\textbackslash{}nObservation'', ``\textbackslash{}nThought''] \\
        max\_tokens & 512   \\ \bottomrule
    \end{tabular}

\end{table}

Table \ref{app_tab:hyper_infer} presents the parameters for invoking the OpenAI API.

\subsection{Experimental Results on Spider 2.0-lite}\label{app_sec:spider2_results}

\begin{table}[ht]
\centering
\caption{Statistics of Spider 2.0-lite (SQLite and Snowflake)}
\label{app_tab:spider2_stats}
\scalebox{1}{
\begin{tabular}{lccccc}
    \toprule
    \textbf{Subtask} & \textbf{\#Item} & \textbf{\#DB} & \textbf{Tb/DB} & \textbf{Col/Tb} & \textbf{FK/DB} \\ \midrule
    SQLite & 135 & 30 & 13.8 & 7.0 & 5.9 \\
    Snowflake & 198 & 51 & 136.6 & 21.8 & 0.0 \\ \bottomrule
\end{tabular}
}
\end{table}

\begin{table}[ht]
\centering
\caption{Results of Spider 2.0-lite (SQLite and Snowflake)}
\label{app_tab:spider2_results}
\scalebox{1}{
\begin{tabular}{lcccc}
    \toprule
    \multicolumn{1}{c}{\multirow{2}{*}{\textbf{Method}}} & \multicolumn{2}{c}{\textbf{SQLite}} & \multicolumn{2}{c}{\textbf{Snowflake}} \\ \cline{2-5} 
    \multicolumn{1}{c}{} & \textbf{EX} & \textbf{\#Token} & \textbf{EX} & \textbf{\#Token} \\ \midrule
    DIN-SQL & 1.50 & 8,334.5 & 1.08 & 59,382.8 \\
    DAIL-SQL (0-shot) & 4.55 & 2,744.4 & 3.45 & 26,052.2 \\
    Ours (2-shot) & \textbf{12.69} & 7,424.6 & \textbf{3.89} & 17,531.8 \\ \bottomrule
\end{tabular}
}
\end{table}

Spider 2.0 \cite{Lei-Fangyu-2024-ArXiv-Spider2} is a sophisticated text-to-SQL framework designed to handle complex queries across various database systems, encompassing diverse SQL dialects and operations. It includes three benchmarks: Spider 2.0, Spider 2.0-snow, and Spider 2.0-lite, where Spider 2.0-lite involves three database systems-SQLite, Snowflake, and BigQuery.
Since these benchmarks focus on various dialects across different databases, which is not the focus of this paper, and considering the cost associated with using Google BigQuery database, we only report the experimental results on the SQLite subset and Snowflake databases subset in Spider 2.0-lite benchmark.

Statistical details in Table \ref{app_tab:spider2_stats} show major differences between SQLite and Snowflake settings. Snowflake has about 10 times more tables per database than SQLite (136.6 vs 13.8) and approximately 30 times more columns per database.

Experimental results, summarized in Table \ref{app_tab:spider2_results}, demonstrate that our method significantly outperforms the baseline. 
On SQLite, our method achieves an 8.1-point improvement over DAIL-SQL. While DAIL-SQL uses the entire DB schema as prompt, Spider 2.0's lack of column descriptions and relatively small schema size (average 96 columns per DB) reduces DAIL-SQL's burden. Our method consumes more tokens due to multi-turn interactions and the inclusion of 2-shot exemplars.

As Spider 2.0 incorporates complex elements such as Common Table Expressions, grouped aggregations, and SQL dialect variations, the primary performance bottleneck lies in the reasoning capabilities of LLMs rather than challenges associated with wide tables.

\subsection{Experimental Results with Open-source LLMs}\label{app_sec:open_llm}

\begin{table}[ht]
    \centering
    \caption{Results on Spider-Dev and its variants with Open-source LLMs}
    \label{app_tab:results_spider_open_source}
    \scalebox{1}{
    
    \begin{tabular}{lcccc}
    \toprule
    \multicolumn{1}{c}{\multirow{2}{*}{\textbf{Method}}} & \multicolumn{2}{c}{\textbf{Spider-Dev}} & \multicolumn{2}{c}{\textbf{Spider-DK}} \\ \cline{2-5} 
    \multicolumn{1}{c}{} & \textbf{EM} & \textbf{EX} & \textbf{EM} & \textbf{EX} \\ \midrule
    GPT-4o & 52.0 & 84.8 & 41.5 & 75.5 \\
    Meta-Llama-3.1-8B-Instruct & 21.4 & 40.2 & 21.9 & 46.4 \\
    Ministral-8B-Instruct-2410 & 28.1 & 51.4 & 28.6 & 50.5 \\ \midrule \midrule
    \multicolumn{1}{c}{\textbf{}} & \multicolumn{2}{c}{\textbf{Spider-Syn}} & \multicolumn{2}{c}{\textbf{Spider-Realistic}} \\ \midrule
    GPT-4o & 44.9 & 78.7 & 55.7 & 81.3 \\
    Meta-Llama-3.1-8B-Instruct & 11.5 & 25.4 & 32.1 & 54.3 \\
    Ministral-8B-Instruct-2410 & 27.1 & 49.9 & 34.4 & 55.9 \\ \bottomrule
    \end{tabular}

}
\end{table}

\begin{table}[ht]
    \centering
    \caption{Results on BIRD-Dev and BIRD Mini-Dev (SQLite) with Open-source LLMs}
    \label{app_tab:results_bird_open_source}
    \scalebox{1}{
    
    \begin{tabular}{lccc}
    \toprule
    \multicolumn{1}{c}{\multirow{2}{*}{\textbf{Method}}} & \textbf{BIRD-Dev} & \multicolumn{2}{c}{\textbf{Mini-Dev}} \\ \cline{2-4} 
    \multicolumn{1}{c}{} & \textbf{EX} & \textbf{EX} & \textbf{Soft F1} \\ \midrule
    GPT-4o & 54.56 & 46.60 & 50.75 \\
    Meta-Llama-3.1-8B-Instruct & 30.96 & 20.40 & 25.94 \\
    Ministral-8B-Instruct-2410 & 28.49 & 20.80 & 23.48 \\ \bottomrule
    \end{tabular}
    }
\end{table}

We conducted experiments using two open-source LLMs, \texttt{Meta -Llama-3.1-8B-Instruct} and \texttt{Ministral-8B-Instruct-2410}, with results shown in Tables \ref{app_tab:results_spider_open_source} and \ref{app_tab:results_bird_open_source}. The results indicate that there remains a considerable performance gap between open-source LLMs and closed-source LLMs.

\section{GenAI Usage Disclosure}
This paper used the web-based version of ChatGPT (accessed via OpenAI's website) solely for grammar corrections and text polishing of the manuscript. No other generative AI tools were used in the research, data collection, analysis, or writing of this paper.

\bibliographystyle{ACM-Reference-Format}

\end{document}